\documentclass[a4paper,fleqn,]{cas-dc}

\usepackage[numbers]{natbib}
\usepackage{graphicx,color,booktabs,multirow,multicol,amsfonts,textcomp,changepage,epstopdf,psfrag,diagbox,enumerate,colortbl,ragged2e,xcolor,times,tikz,algorithmicx,algpseudocode,cases,stfloats,bm}
\usepackage{float,booktabs}
\usepackage{pifont}

\usepackage[tight,footnotesize,center]{subfigure}

\usepackage{lineno,hyperref,algorithm}

\definecolor{cblue}{rgb}{0.21, 0.49, 0.74}

\hypersetup{
	final,
	breaklinks,
	colorlinks,
	linkcolor={red},   
	citecolor={cblue},
	urlcolor={Mahogany}     
}

\def\tsc#1{\csdef{#1}{\textsc{\lowercase{#1}}\xspace}}
\tsc{WGM}
\tsc{QE}
\tsc{EP}
\tsc{PMS}
\tsc{BEC}
\tsc{DE}

\begin{document}
\let\WriteBookmarks\relax
\def\floatpagepagefraction{1}
\def\textpagefraction{.001}
\hypersetup{pageanchor=false}
\shorttitle{}

\shortauthors{X. Zhang et~al.}
\title [mode = title]{InScope: A New Real-world 3D Infrastructure-side Collaborative Perception Dataset for Open Traffic Scenarios}      
\author[1]{Xiaofei Zhang}
\ead{zhangxf87@mail2.sysu.edu.cn}

\credit{Conceptualization, Methodology, Software, Data curation, and Writing - original draft}


\author[1]{Yining Li}
\ead{liyn69@mail2.sysu.edu.cn}
\credit{Data curation, and Methodology}

\author[1]{Jinping Wang}
\ead{wangjp29@mail2.sysu.edu.cn}
\credit{Data curation, Methodology, and Writing - original draft}

\author[1]{Xiangyi Qin}
\ead{qinxy36@mail2.sysu.edu.cn}
\credit{Data curation, and Methodology}

\author[1]{Ying Shen}
\ead{sheny76@mail.sysu.edu.cn}
\credit{Methodology, and Supervision}

\author[1]{Zhengping Fan}
\ead{fanzhp@mail.sysu.edu.cn}
\credit{Investigation, Supervision, and Formal analysis}

\affiliation[1]{organization={School of Intelligent Systems Engineering, Sun Yat-sen University},
    city={Shenzhen},
    postcode={518107}, 
    country={China}}

\author[2]{Xiaojun Tan}
\cormark[1]
\ead{tanxj@mail.sysu.edu.cn}
\credit{Validation, Funding acquisition, and Supervision}
\cortext[cor1]{Corresponding author}

\affiliation[2]{organization={School of Intelligent Systems Engineering, Sun Yat-sen University, Shenzhen, 518107 \& Southern Marine Science and Engineering Guangdong Laboratory (Zhuhai), Zhuhai, 519082, China}}


\newcommand{\eg}[0]{\textit{e.g.,}}
\newcommand{\ie}[0]{\textit{i.e.,}}
\newcommand{\name}[0]{\textit{InScope}}

\begin{abstract}
Perception systems of autonomous vehicles are susceptible to occlusion, especially when examined from a vehicle-centric perspective. Such occlusion can lead to overlooked object detections, \eg larger vehicles such as trucks or buses may create blind spots where cyclists or pedestrians could be obscured, accentuating the safety concerns associated with such perception system limitations. To mitigate these challenges, the vehicle-to-everything (V2X) paradigm suggests employing an infrastructure-side perception system (IPS) to complement autonomous vehicles with a broader perceptual scope. Nevertheless, the scarcity of real-world 3D infrastructure-side datasets constrains the advancement of V2X technologies. To bridge these gaps, this paper introduces a new 3D \textbf{in}frastructure-\textbf{s}ide \textbf{co}llaborative \textbf{pe}rception dataset, abbreviated as \name. Notably, InScope is the first dataset dedicated to addressing occlusion challenges by strategically deploying multiple-position Light Detection and Ranging (LiDAR) systems on the infrastructure side. Specifically, InScope encapsulates a 20-day capture duration with  303 tracking trajectories and 187,787 3D bounding boxes annotated by experts. Through analysis of benchmarks, four different benchmarks are presented for open traffic scenarios, including collaborative 3D object detection, multisource data fusion, data domain transfer, and 3D multiobject tracking tasks. Additionally, a new metric is designed to quantify the impact of occlusion, facilitating the evaluation of detection degradation ratios among various algorithms. The Experimental findings showcase the enhanced performance of leveraging InScope to assist in detecting and tracking 3D multiobjects in real-world scenarios, particularly in tracking obscured, small, and distant objects. The dataset and benchmarks are available at \url{https://github.com/xf-zh/InScope}.

\end{abstract}

\maketitle{}
\section{Introduction}
\label{sec:intro}
\begin{figure*}
  \centering{
  \includegraphics[scale=1.5]{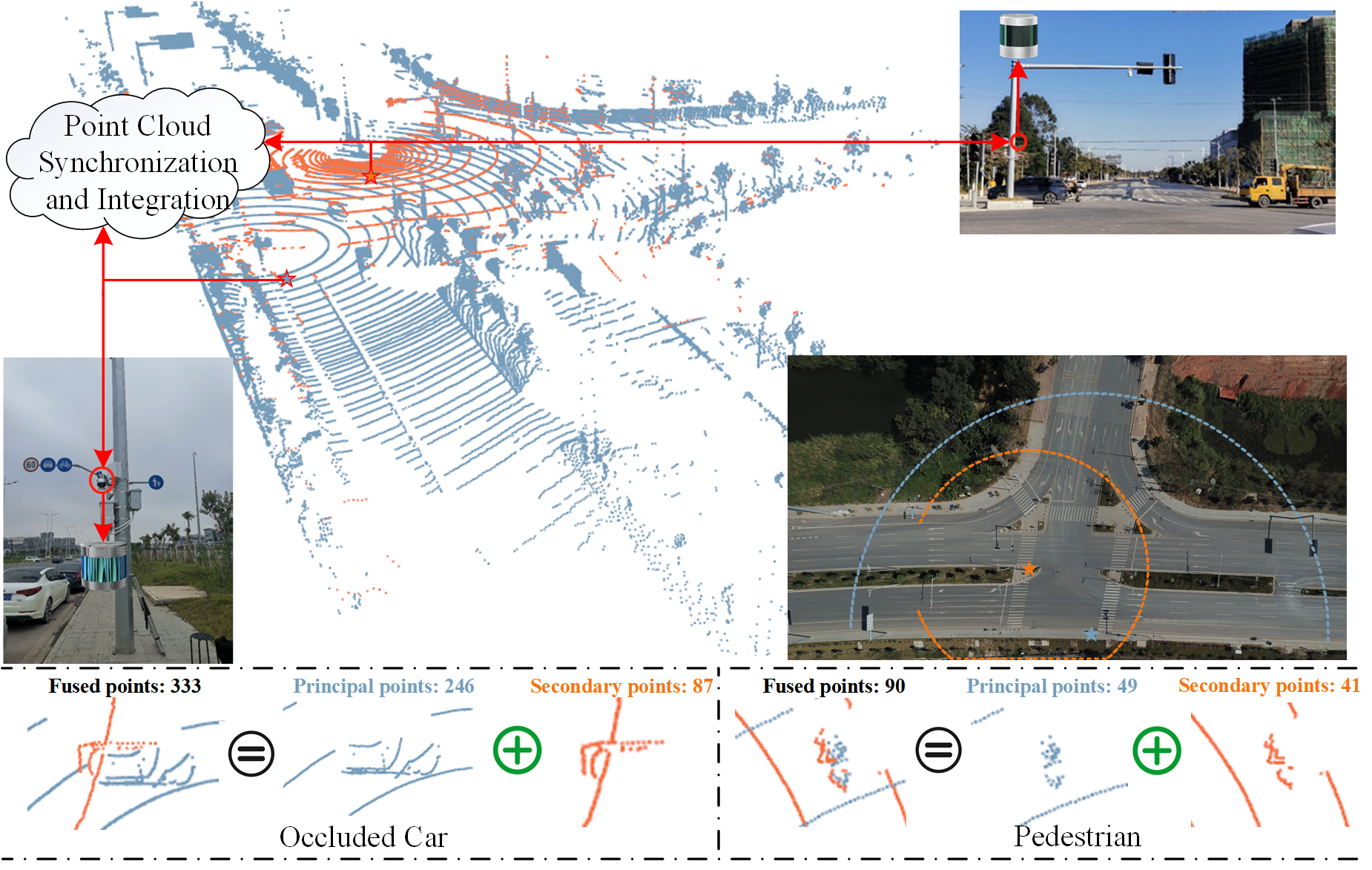}}
  \caption{\textbf{InScope overview.} The blue and yellow dashed lines indicate the detection areas of the principal and secondary LiDAR systems, respectively, installed on the points of the blue and yellow pentagrams. Specific examples of fusion points are given at the bottom.}
  \label{fig:dataset1}
\end{figure*}

Recently, autonomous driving has become a prominent area of research \cite{wang2023multi,fernandes2021point}. Numerous high-quality datasets for bolstering vehicle-side perception system (VPS) research within the autonomous driving community have been released, yielding compelling results \cite{10373157,Wu102370}. Nevertheless, the restricted field of view of the VPS frequently results in a short-range, incomplete perception of scene information \cite{chen2023transiff}. In response to this challenge, the infrastructure-side perception system (IPS) has been proposed as a resolution to supplement the restrictions of the VPS. IPSs are commonly integrated into infrastructure facilities (refer to Figure \ref{fig:dataset1}), facilitating an expansive viewpoint and long-range perception capabilities \cite{jia2023monouni,Huang101834} and thereby supporting the vehicle-to-everything (V2X) \cite{chen2023transiff}.  

To expedite V2X research, the DAIR-V2X \cite{yu2022dair} and V2X-Seq \cite{yu2023v2x} datasets collaboratively integrate infrastructure- and vehicle-side perspectives to enhance the safety of autonomous driving. Despite this progress, their reliance solely on the infrastructure-side viewpoint leaves them susceptible to occlusions caused by larger vehicles, resulting in short-term blind spots. These blind spots pose a potential risk of missed detections, compromising the reliability and security of V2X systems, especially at intersections \cite{buchholz2021handling}.

To overcome the deficiencies, this paper introduces a novel \textbf{in}frastructure-\textbf{s}ide \textbf{co}llaborative \textbf{pe}rception system named \textbf{\name}, which can cover potential blind spots in the perceptual field via the cooperation of multi-position LiDARs. Figure~\ref{fig:dataset1} illustrates the capture process and motivation for InScope, emphasizing infrastructure equipped with sensors at an elevated position and providing a broader site-of-view. The InScope capturing system integrates two LiDARs positioned at distinct locations. The primary LiDAR is designed to capture essential perception data, whereas the secondary LiDAR complements this role by capturing information that pertains to addressing blind spots. The data subsets gathered by these LiDAR systems are identified as InScope-Pri and InScope-Sec, respectively.

The lower segment of Figure \ref{fig:dataset1} illustrates two instances of point clouds representing cars and pedestrians. While points from the principal or secondary LiDARs individually contribute to point segments, neither captures the complete boundary. In contrast, fused points combine all accessible cloud points sourced from both LiDAR, providing improved object boundary delineation for cars or pedestrians, thereby enhancing the comprehensive perception of adjacent traffic surroundings. To our knowledge, the InScope is the first large-scale dataset focusing on anti-occlusion perception for infrastructure-to-infrastructure (I2I) collaborative perception. It is an ideal solution for achieving blind-spot-free perception in V2X scenarios.

In addition, the DAIR-V2X dataset only supplies the 3D object detection benchmark without providing trajectory annotation. Based on the DAIR-V2X dataset, the V2X-Seq dataset provides additional annotations to achieve trajectory tracking and forecasting. 

Compared with the DAIR-V2X and V2X-Seq datasets, the proposed InScope dataset can meet four benchmarks simultaneously without additional annotation: 3D object detection, multisource data fusion, data-domain transfer, and 3D multiobject tracking. Specifically, \textbf{3D object detection} benchmark mainly counts the 3D information of traffic participants (such as cars, trucks, cyclists, pedestrians, \textit{etc.}), significantly expanding the perception field of autonomous vehicles. The \textbf{multisource data fusion} benchmark can evaluate the robustness of multisource data with different paradigms and rigorously assess the impact of fusion mechanisms on perception performance. \textbf{Data-domain transfer} realizes development and verification of unsupervised learning method. It also facilitates the application of SOTA 3D detectors, which were initially conceived for vehicle-side datasets. The \textbf{3D multiobject tracking} benchmark extends the capabilities of 3D object detection by tracking object movement over time, where the obscured, small, and distant objects can be tracked completely within V2X scenarios, providing a dynamic and comprehensive understanding of traffic surroundings.\par

\begin{table*}
    \caption{\textbf{Dataset comparison.} 2D and 3D represent the image and point cloud, respectively. Ego: Egocentric vehicle. V2V: Vehicle cooperative system, Infra: Infrastructure, I2I: Infrastructure cooperative system, V2X: Vehicle-infrastructure cooperative system. ``Real/Sim.'' indicates that the datasets are collected in the real or simulated world. 
}
    \centering
    \footnotesize
    \setlength{\tabcolsep}{4pt}
    
     \resizebox{1\textwidth}{!}{
    \begin{tabular}{cccccccccc}
    \toprule
    \multirow{2}{*}{Dataset}
    
    
    &   \multirow{2}{*}{View}  & \multirow{2}{*}{Real/Sim}
     & \multirow{2}{*}{Year}
      & Point& 3D  & Trajectory & Blind-Spot & Occlusion  & \multirow{2}{*}{Benchmark}\\

&&&&Clouds& Boxes &Annotation & Awareness  & Quantitative  & 
\\

    
    \midrule
    KITTI \cite{geiger2012we}                   & Ego & Real  & 2012  & 15k   & 200k  & \ding{51} & \ding{55}  & \ding{55} & 5 \\
    ApolloScape \cite{huang2018apolloscape}     & Ego & Real  & 2018  & -     & 70k   & \ding{51} & \ding{55}  & \ding{55} & 5 \\
    nuScenes \cite{caesar2020nuscenes}          & Ego & Real  & 2020  & 400k  & 1.4M  & \ding{51} & \ding{55}  & \ding{55} & 2 \\
    Waymo \cite{sun2020scalability}             & Ego & Real  & 2020  & 200k  & 12M   & \ding{51} & \ding{55}  & \ding{55} & 2 \\
    ONCE  \cite{mao2021one}                     & Ego & Real  & 2021  & 1M    & 417k  & \ding{55} & \ding{55}  & \ding{55} & 4 \\
    V2V4Real \cite{xu2023v2v4real}              &      V2V       & Real  & 2023  & 20k   & 240K  & \ding{51} & \ding{51}  & \ding{55} & 3 \\
    \midrule
    V2X-Sim \cite{li2022v2x}                    &      V2X       & Sim.   & 2022  & 10k   & 26.6k & \ding{51} & \ding{51}  & \ding{55} & 4 \\
    V2X-Seq/Perception \cite{yu2023v2x}         &      V2X       & Real  & 2023  & 15k   & 12k   & \ding{51} & \ding{51}  & \ding{55} & 2 \\
    \midrule
    NGSIM \cite{NGSIM}                          & Infra & Sim.   & 2016  & -     & -     & \ding{51} & \ding{55}  & \ding{55} & 7 \\
    WIBAM \cite{WIBAM}                          & Infra & Real  & 2021  & -    & -     & \ding{55} & \ding{55}  & \ding{55} & 1 \\
    DAIR-V2X-I \cite{yu2022dair}                & Infra  & Real  & 2022  & 10k   & 493k  & \ding{55} & \ding{55}  & \ding{55} & 1 \\
    RCooper (Corridor) \cite{hao2024rcooper}      &I2I     & Real  & 2024  & 11k    & 53k   & \ding{51} & \ding{51}  & \ding{55} & 2 \\
    RCooper (Intersection) \cite{hao2024rcooper}  &I2I    & Real  & 2024  & 2k    & 43k   & \ding{51} & \ding{51}  & \ding{55} & 2\\
    \midrule
    \textbf{InScope} (Intersection)                     & I2I         & Real  & 2024  & 21k   & 188k  & \ding{51} & \ding{51}  & \ding{51} & 4\\
    \bottomrule
    \end{tabular}
    }
  \label{tab:dataset}
\end{table*}


The main contributions of this study can be summarized as follows:
\begin{itemize}
\item A large-scale dataset (InScope) featuring multi-position LiDARs in a real-world setting is presented in this paper, specifically developed to address the current research gap concerning occlusion challenges within the I2I perception systems in open traffic scenarios.\par

\item The InScope dataset is devoid of sensitive information and incorporates trajectory annotations alongside 188K 3D bounding boxes. This paper provides four benchmarks to thoroughly introduce the dataset, aiming to make a valuable contribution to the V2X research community.\par

\item Another metric ($\xi_D$) within the I2I setting is further formulated to evaluate the anti-occlusion capabilities of the I2I dataset systematically. This metric gauges the degradation ratios in detection performance between scenarios featuring only one LiDAR sensor and scenes with multiple LiDARs.\par

\item Several commonly employed baseline experiments have been reproduced on InScope. The comprehensive experimental findings affirm the efficacy of utilizing InScope to explore model anti-occlusion across different real-world datasets, thereby establishing robust groundwork for forthcoming research endeavors to address this challenge.\par
\end{itemize}

\section{Related Work}
\label{related work} 
\subsection{Vehicle-side Datasets}
Public vehicle-side datasets involving various vision tasks have significantly promoted the process of autonomous driving \cite{caesar2020nuscenes,sun2020scalability,Wang102247}. Moreover, due to the inherent limitations of unimodal autonomous perception, multimodal cooperative perception has become a longstanding focus in the V2X field \cite{cai20223d,guo2023joint,su2024makes,yu2024flow}. Typical multimodal perception systems involve single-ego camera-camera or camera-LiDAR systems, as summarized in Table \ref{tab:dataset}. The most relevant  Karlsruhe Institute of Technology and Toyota Technological Institute (KITTI) dataset released in 2012 labelled 200k 3D object bounding boxes \cite{geiger2012we}. It can be used for stereo, optical flow, visual odometry / SLAM, and 3D object detection. Afterwards, a large-scale and real-world dataset named the One millioN sCenEs (ONCE) was released with more than 417k annotations \cite{mao2021one}. To support more complex and intelligent perception tasks, nuScenes \cite{caesar2020nuscenes} and Waymo \cite{sun2020scalability} datasets were introduced to implement object detection, multiobject tracking, autonomous driving system validation and testing, motion planning, and other tasks. Later, the V2V4Real dataset was used to improve the long-term perception capability of autonomous vehicles by introducing a multi-vehicle cooperative perception system in real scenes \cite{xu2023v2v4real}.

Despite recent progress in ego vehicle-side perception, it often suffers from perceptual degradation phenomena in long-range or occluded areas \cite{10443542,Chen_2023_ICCV}.

\subsection{Infrastructure-side and V2X Datasets}
Single vehicle-side perception significantly impedes the ability of autonomous vehicles to make more intelligent decisions \cite{yu2022dair}. Recently, infrastructure-side and vehicle-infrastructure cooperative (V2I) datasets have received increasing attention as a supplement to V2X. Several publicly available datasets are accessible for infrastructure-side and V2I applications, such as NGSIM \cite{NGSIM}, WIBAM \cite{WIBAM}, DAIR-V2X-I \cite{yu2022dair}, and V2X (V2X-Sim \cite{li2022v2x}, DAIR-V2X-C \cite{yu2022dair}, and V2X-Seq \cite{yu2023v2x}). Specifically, for infrastructure-side datasets, NGSIM mainly counts vehicle motion trajectories, whereas WIBAM and DAIR-V2X-I count the 3D bounding boxes for developing 3D object detection methods. For the V2X datasets, V2X-Sim annotates the 3D bounding boxes and motion trajectories. However, its data are often sourced from simulations rather than real-world scenarios, causing a large gap between theory and practical implementation. Later, the DAIR-V2X-C and V2X-Seq datasets were proposed to reflect more complexity and irregularity of the real world and can serve as a blind spot supplement (BSS) for the VPS based on the IPS. However, the effectiveness of the DAIR-V2X-C and V2X-Seq datasets may be compromised in scenarios where occlusion obstructs the acquisition of infrastructure-side information, and distant objects are sparsely perceived. These issues can result in catastrophic accidents and present challenges that are not readily addressed by existing methods.

In addition, frequent occlusions create significant hazards in real-world traffic scenarios. These factors complicate monitoring, navigation, and safety, increasing the risk of accidents. To reduce infrastructure-side blind spots, the RCooper dataset is proposed to provide information compensation through infrastructure-side LiDAR located at different locations. The RCooper dataset contains two scenes, i.e., the intersection and the corridor scene. Although the intersection scene in the RCooper dataset can serve infrastructure-to-infrastructure (I2I) BSS, the frames of the point cloud and the number of annotations are limited. In contrast, the corridor scene in the RCooper dataset mainly provides more extended temporal information for the continuous tracking of objects, which makes it challenging to meet the I2I BBS task.

Generally, subjective common sense is used to assess the anti-occlusion capability of a method, with higher accuracy in object detection often perceived as indicative of more robust resistance to interference. However, the lack of a specialized dataset for quantitatively evaluating anti-occlusion capabilities means that researchers and developers might not have the necessary tools to rigorously test and refine their methods under occluded conditions.

As a solution, a new infrastructure-to-infrastructure collaborative BSS system named InScope is developed, offering an anti-occlusion test benchmark and more comprehensive infrastructure-side scene information. It can help provide continuous, stable, and accurate perception information for autonomous vehicles, especially for those areas with traffic occlusions.

\begin{table}
 \caption{Equipment used in the InScope and the key indicators of the sensors. The positions of the principal LiDAR and secondary LiDARs are shown in Figure \ref{fig:dataset1}.}
  \centering  
  \footnotesize
    \begin{tabular}{lp{5.4 cm}}
        \toprule
    Sensors & Details \\
        \midrule
    Principal LiDAR & 80-beam, 10Hz capture frequency, vertical angular resolution of 0.1°, $\le$ 230$m$ range, +90° $\sim$ +275° horizontal FOV. \\
    Secondary LiDAR & 32-beam, 10Hz capture frequency, vertical angular resolution of 1.33°, $\le$ 150$m$ range, +15° $\sim$ +305° horizontal FOV. \\
        \bottomrule
  \end{tabular}
\label{tab:lidars}
\end{table}

\section{InScope Dataset}
\label{SYSTEM}
There are three datasets discussed in this paper, including the InScope-Pri, InScope-Sec, and InScope datasets. Specifically, InScope-Pri and InScope-Sec are two datasets captured from single principal and secondary LiDARs, respectively, featuring partial and potential occlusions. The InScope dataset is composed of both the InScope-Pri and InScope-Sec datasets, offering more comprehensive data from fused LiDARs.

The situated position and perception field of view of the InScope are shown in Figure \ref{fig:dataset1}. Notably, the blue and yellow pentagrams in the lower right picture represent the locations of the principal and secondary LiDARs, respectively. The blue and yellow dashed lines represent the perception ranges of the principal and secondary LiDARs, respectively. The detailed specifications of the equipment parameters utilized in InScope are described in Table \ref{tab:lidars}. 

The system setup will be introduced in Section \ref{setup}. Section \ref{annotation} describes the collection and annotation manner of the InScope dataset.

\subsection{System Setup}
\label{setup}
\subsubsection{Sensors} The principal LiDAR captures the primary data, and the secondary LiDAR captures the supplementary data. Based on this idea, a high-precision 80-beam LiDAR with 10 Hz sampling is adopted as the principal LiDAR in the InScope. Its maximum detection range is 230 $m$. Since the west of the T-intersection is industrial, where monitoring is not allowed, the horizontal view of the principal LiDAR is set to $[90^{\circ}, 275^{\circ}]$. In busy scenes, trucks obstructing intersection can occlude other objects. This lost information could cause erroneous judgements of vehicle-infrastructure coordination and perception errors for autonomous vehicles. To minimize occlusion interference, a secondary LiDAR compensates for blind-spot coverage, specifically within the intersection. Here, a low-cost 32-beam unit with a 150-$m$ range is used for temporary supplementation. The secondary LiDAR faces south, and its detection area is set to $[15^{\circ}, 305^{\circ}]$.\par

\subsubsection{Temporal Consistency}
\begin{figure}
  \centering{
  \includegraphics[scale=0.7]{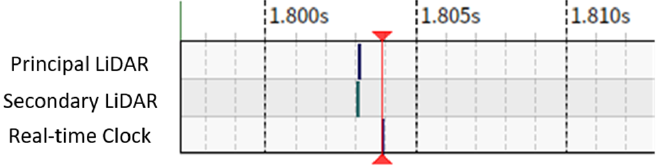}}
  \caption{The temporal consistency among two LiDARs and the computer clock.}
  \label{fig:Temporal}
\end{figure}

 Recent studies \cite{yu2022dair,xu2023v2v4real} have proven that a temporal variance from 10 ms (DAIR-V2X) to 50 ms (V2V4Real) between multiple LiDARs can ensure the accuracy and effectiveness of data fusion. Figure \ref{fig:Temporal} indicates that the time error between two LiDARs in the InScope meets the requirements (at the microsecond level). The time information was recorded from the \emph{ROS} bags.

\subsubsection{Calibration}
The InScope dataset contains three Cartesian coordinates: principal LiDAR coordinate, secondary LiDAR coordinate, and world coordinate. For the principal LiDAR coordinate, the X-axis and Y-axis are oriented towards the east and north, respectively. For the secondary LiDAR coordinate, the X-axis and Y-axis are oriented towards the south and east, respectively. The InScope uses principal LiDAR coordinate as unified coordinates. More than 20 calibration points were collected to obtain the secondary-to-principal transform matrix, and their spatial location in three Cartesian coordinates was recorded. From these data, the secondary-to-world and world-to-principal transformation matrix can be obtained. Finally, the above two transform matrixes are multiplied to determine the critical secondary-to-principal transform matrix \cite{Qiu101806}. 

Considering the interference caused by remote and sparse point clouds,
the area of interest of the InScope dataset $\left[x_{min}, y_{min}, z_{min}, x_{max}, y_{max}, z_{max}\right]$ is set to a rectangular area $\left[-5.0, -75.2, -5.0, 67.0, 75.2, 3.0\right]$.

\subsection{Data Collection, Annotation, and Statistics}
\label{annotation}
For regulatory reasons, the detailed installation site of InScope in Guangdong Province, China, will remain undisclosed. Since InScope is a pure point cloud dataset, it does not contain sensitive information other than map data and will not violate personal rights. \par

\subsubsection{Collection and Annotation} After collecting the raw data, 87 sequences were manually selected in various weather environments, with a period of 20 days and 21,317 frames of point cloud data. The Experts then manually annotated the dataset using a redeveloped high-precision annotation tool. They applied category, 3D bounding boxes (including x, y, z, width, length, height, and orientation), and tracking IDs to each frame through rigorous full supervision, ensuring labelling quality. For each 3D bounding box in the InScope dataset annotation, if no corresponding 3D box in the InScope-Pri or InScope-Sec dataset annotations--matching in both location and category--is identified, the boxes of the InScope dataset are integrated into the InScope-Pri or InScope-Sec dataset annotations. This process facilitates the creation of I2I cooperative annotations for the InScope dataset. These cooperative annotations are manually supervised and refined to ensure the generation of more precise annotations.

\begin{figure}
  \centering{\includegraphics[scale=0.84]{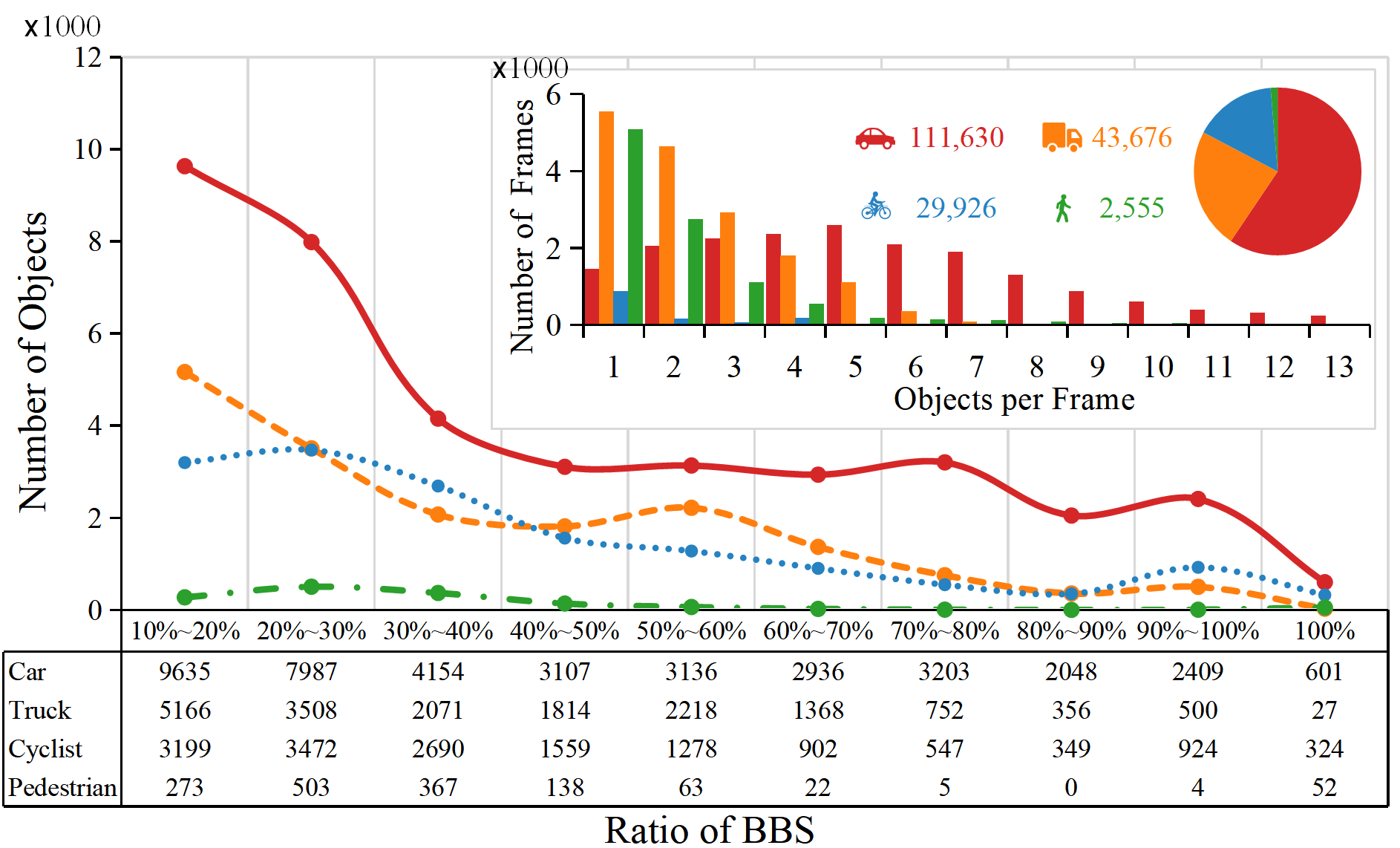}}
  \caption{Distribution of objects with different ratios of BBS. The subfigure reports the distribution of object counts per frame.}
  \label{fig:dataset3}
\end{figure}

As mentioned above, the InScope-Sec dataset is mainly responsible for providing blind spots supplement data for the InScope dataset. In Figure \ref{fig:dataset3}, the ratio of InScope-Sec point cloud and InScope point cloud in each object is shown, defined as the ``Ratio of BBS''. It can be found that the InScope-Sec dataset can provide at least 50\% compensation data for nearly one out of three of the objects, proving that the InScope dataset is suitable for tasks of collaborative perception. In addition, the average number of various types of objects in each frame is counted in the sub-figure of Figure \ref{fig:dataset3}. It can be seen that there are eight objects per frame on average, and the InScope replicates real traffic scenarios authentically. The annotated ground truth and expert validation establish the InScope dataset as a benchmark for developing and evaluating next-generation multisource perception systems. 

Finally, Figure \ref{fig:dataset3} reports the total number of various classes of objects in the InScope dataset, including 111,630 cars, 43,676 trucks, 29,926 cyclists, and 2,555 pedestrians. The scene is located in an industrial area, and the proportion of trucks and cars accounts for 82.70\% of the total number of objects.

\begin{figure}
  \centering{\includegraphics[scale=1]{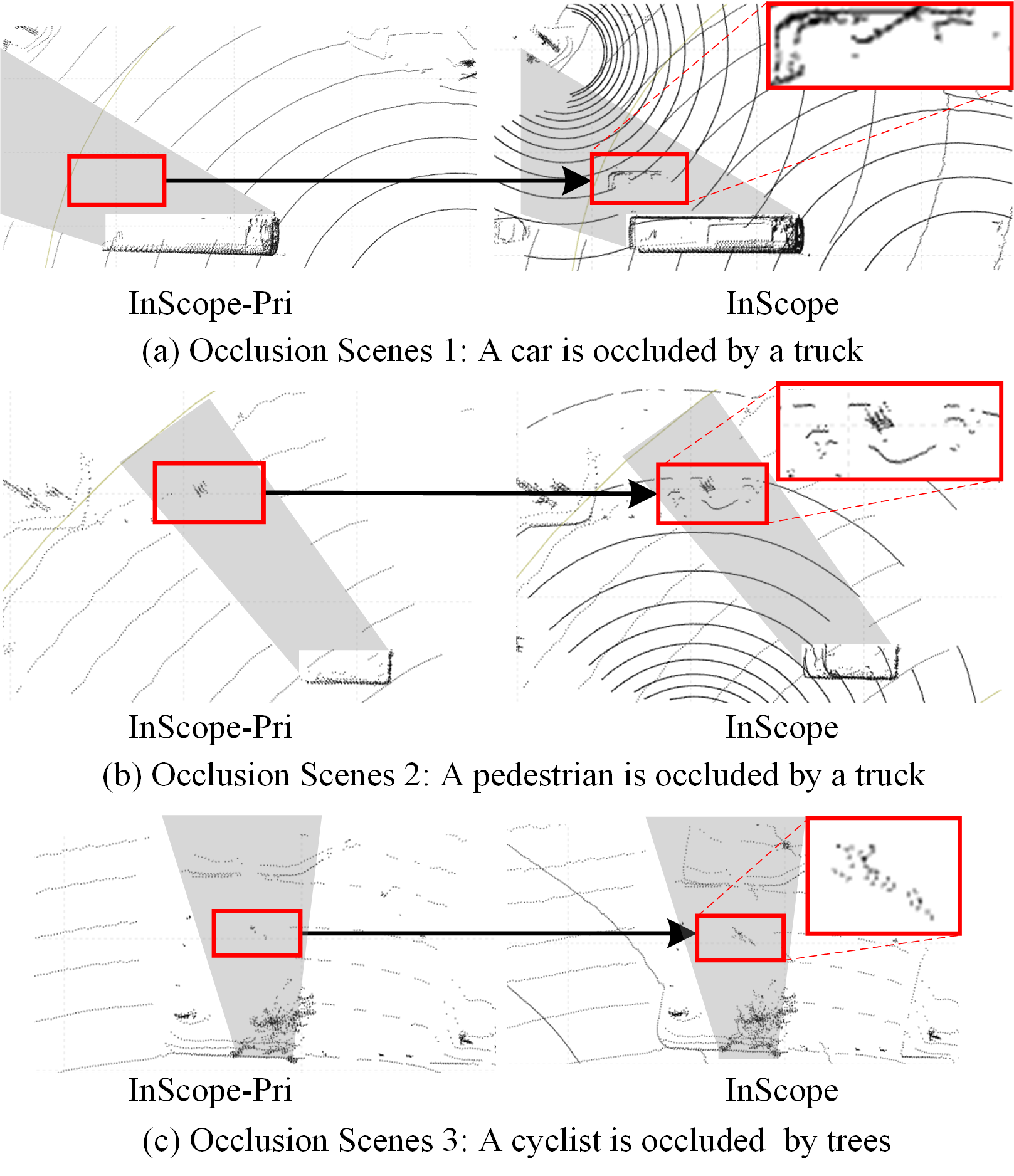}}
  \caption{Real-world occlusion examples from the InScope-Pri and InScope datasets are depicted. Red boxes indicate objects occluded in blind-spots (grey regions), which cannot be detected by the InScope-Pri dataset.}
    \label{fig:dataset2}
\end{figure}

\subsubsection{Complex Occlusion Scenes} The InScope dataset includes various occlusion scenarios crafted to challenge and assess the resilience of methods. These include (1) cars, cyclists, and pedestrians occluded by trucks; (2) cyclists and pedestrians occluded by trucks or cars; and (3) trucks, cyclists, and pedestrians partially occluded by trees or other surroundings. Innovation can be fostered by identifying specific weaknesses and encouraging targeted improvements in the method design process.

Figure \ref{fig:dataset2} shows specific real-world examples of occluded objects in the InScope dataset. Due to occlusion, the grey region denotes the perception gaps from only the InScope-Pri dataset. With the addition of a secondary LiDAR, the InScope dataset successfully perceives the occluded objects marked with red rectangles (such as cars, pedestrians, and cyclists) within the blind spot. These compensation data are critical for continuously tracking objects within a visual blind spot. More importantly, these examples demonstrate the ability of InScope to harvest crucial supplementary information, particularly for objects in the center of an intersection. By mitigating these blind spots, InScope strengthens situational awareness of traffic scenes – a critical capability for advanced applications of autonomous vehicles.

\section{Experiments on the InScope Dataset}
This section records and reports the results of experiments involving the four tasks on the InScope dataset as a baseline for future research. Moreover, the impact of the InScope data characteristics on different benchmarks is also analyzed. Finally, suggestions for future research on the benchmarks are given. The experiments are performed on a Linux server with 3 $\times$ A100 GPUs and 175 GB of RAM and reported on the validation set.

\begin{table*}
  \centering
  \caption{Comparative 3D detection results of different backbones on the InScope-Pri and InScope datasets, with the best performance marked in \textbf{bold}.}
  \footnotesize
    \begin{tabular}{cccccccc}
    \toprule
    Datasets & Methods & \multicolumn{1}{m{6em}<{\centering}}{\ \ Car ↑\newline{}$AP_{40(IoU=0.7)}$} & \multicolumn{1}{m{6em}<{\centering}}{Pedestrian ↑\newline{}$AP_{40(IoU=0.5)}$} & \multicolumn{1}{m{6em}<{\centering}}{Cyclist ↑\newline{}$AP_{40(IoU=0.5)}$} & \multicolumn{1}{m{6em}<{\centering}}{\  Truck ↑\newline{}$AP_{40(IoU=0.7)}$} & $mAP_{40}$ ↑ & FPS ↑\\
    \midrule
    \multirow{4}[0]{*}{InScope-Pri} & PointRCNN & 61.14  & \textbf{61.99} & 48.96  & 88.80  & 65.22  & 4.57  \\
          & Pointpillar & 67.34  & 23.82  & 43.51  & \textbf{91.59} & 56.57  & \textbf{41.32} \\
          & PV-RCNN++ & \textbf{72.59} & 45.26  & \textbf{61.21} & 91.02  & \textbf{67.52} & 13.28  \\
          & CenterPoint & 61.31  & 49.62  & 52.73  & 82.02  & 61.42  & 31.45  \\
      \midrule
    \multirow{4}[0]{*}{InScope} & PointRCNN & 71.75  & 68.13  & 62.91  & 94.50  & 74.32  & 4.58  \\
          & Pointpillar & 78.04  & 35.34  & 58.46  & 95.86  & 66.93  & 24.51  \\
          & PV-RCNN++ & \textbf{80.55} & 53.31  & 70.92  & 95.92  & 75.18  & 14.66  \\
          & CenterPoint & 77.24  & \textbf{70.45} & \textbf{74.74} & \textbf{96.12} & \textbf{79.64} & \textbf{30.49} \\
    \bottomrule
    \end{tabular}
  \label{tab:detction_inscope}
\end{table*}

\subsection{Collaborative 3D Object Detection}

\subsubsection{Metrics} 
\label{sec:metrics}
$\bullet$ The average precision (AP) using 40 recall positions ($AP_{40}$): The popular $AP_{40}$ is adopted to measure the detection performance of 3D object detectors \cite{simonelli2019disentangling}. Following the conventions of KITTI and nuScenes, for larger objects such as cars and trucks, a higher intersection over union (IoU) threshold (IoU=0.7) is used to evaluate the detection performance of the methods rigorously. In contrast, considering that the 3D sizes of cyclists and pedestrians are smaller, a lower IoU threshold (IoU=0.5) is used to evaluate the detection performance. $AP_{40} \in \left[0, 1\right]$.

$\bullet$ The mean $AP_{40}$ ($mAP_{40}$): The average $AP_{40}$ of all classes is adopted \cite{9983516} toevaluate the detection performance of methods more comprehensively. $mAP_{40} \in \left[0, 1\right]$.

$\bullet$ The frames per second (FPS): FPS is adopted to evaluate the computational efficiency of methods. 

$\bullet$ The mean detection performance degradation ratio $\xi_{D}$: This paper designs a new metric to quantitatively evaluate the collaborative perception performance of 3D object detection methods. This metric comprehensively considers the differences between methods in the InScope and InScope-Pri datasets and can objectively and comprehensively evaluate the collaborative perception performance of the methods in infrastructure-side datasets. In this way, $\xi_{D}$ can be defined as:
\begin{equation}
\xi_{D} = 1 - {G}_{mclass} \times {G_{mAP}},
\end{equation}
where ${G}_{mclass}$ and ${G_{mAP}}$ represent the mean performance gap ratio of all classes and the performance degradation ratio of $mAP_{40}$, respectively. ${G_{mAP}} \in \left[0, 1\right]$. The lower the degradation ratio of detection performance, the better the anti-occlusion performance of methods.

For the ${G}_{mclass}$ metric, it can be expressed as:
\begin{equation}
{G}_{mclass} = \frac{1}{c}\sum\limits_c {G_c} = \frac{1}{c}\sum\limits_c {\frac{{AP_{40}{{\left( {InScope_{Pri}} \right)}_c}}}{{AP_{40}{{\left( {InScope} \right)}_c}}}},
\end{equation}
where $AP_{40}(InScope_{Pri})$ and $AP_{40}(InScope)$ represent the AP of each class in the InScope-Pri and InScope datasets. ${c \in \left[ {car,pedestrian,cyclist,truck} \right]}$ represents the object classes. ${G_c}$ represents the detection performance gap ratio of each class on the InScope-Pri and InScope datasets. ${G_{mclass}} \in \left[0, 1\right]$.

The ${G_{mAP}}$ metric can be formulated as:
\begin{equation}
\begin{aligned}
{G_{mAP}} &= \frac{{mA{P_{40}}\left( {InScope_{Pri}} \right)}}{{mA{P_{40}}\left( {InScope} \right)}} \\
&= \frac{{\sum\limits_{c} {A{P_{40}}{{\left( {InScope_{Pri}} \right)}_c}} }}{{\sum\limits_{c}^{} {A{P_{40}}{{\left( {InScope} \right)}_c}} }},
\end{aligned}
\end{equation}
where $mAP_{40}(InScope_{Pri})$ and $mAP_{40}(InScope)$ represent the $mAP_{40}$ in the InScope-Pri and InScope datasets, respectively. ${G_{mAP}} \in \left[0, 1\right]$.

\subsubsection{Baselines} 
3D object detection is critical for both the InScope-Pri and InScope datasets. Additionally, analyzing the anti-occlusion performance of various detector backbones is essential. More details are shown as follows.\par

\begin{figure*}
  \centering{
  \includegraphics[scale=1]{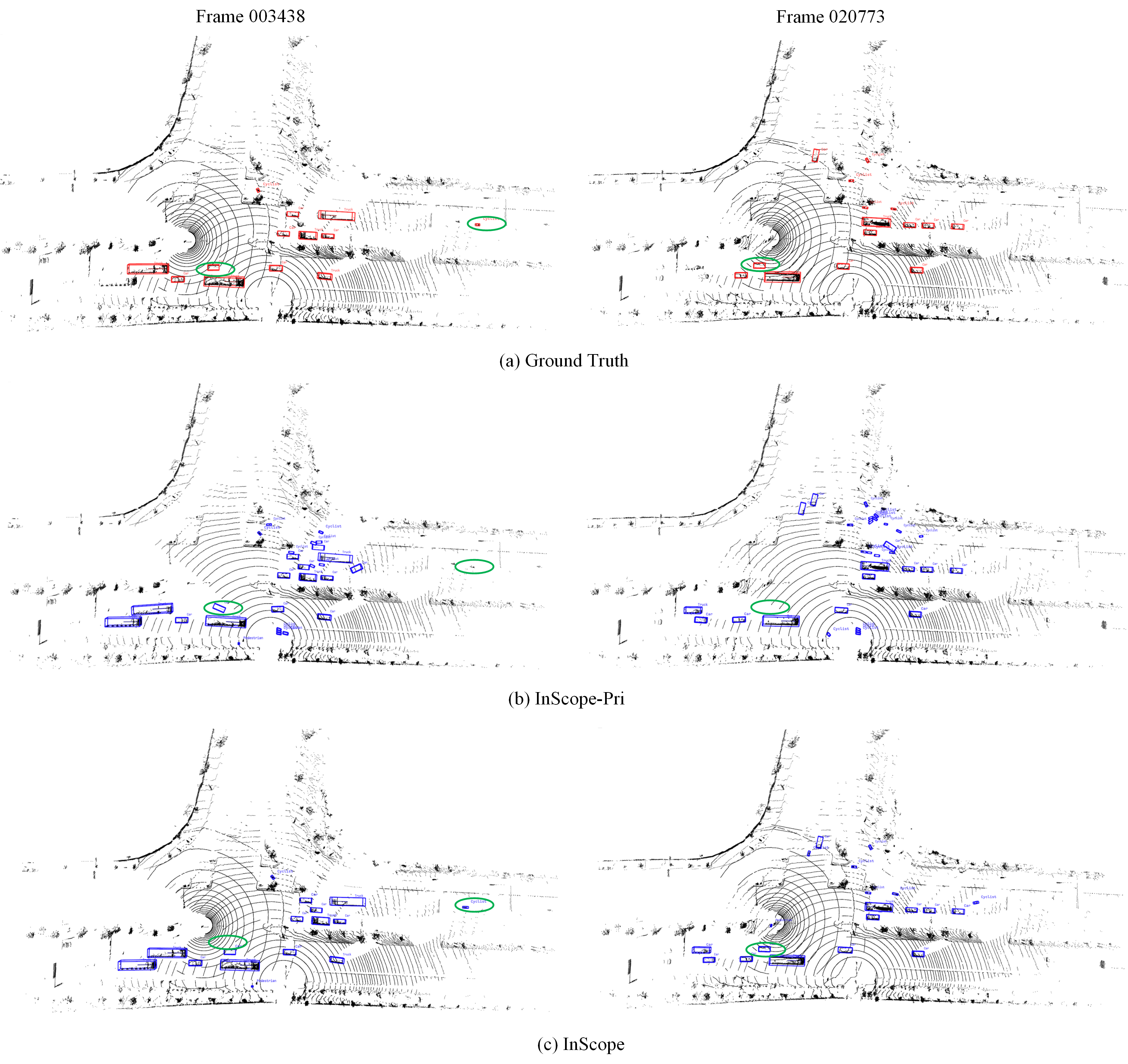}}
  \caption{The detection results of the CenterPoint method on the InScope-Pri and InScoperi datasets. The red and blue bounding boxes represent the ground truth and detection results, respectively. The green circles identify the result differences of CenterPoint based on different datasets.}
  \label{fig:detection_compare}
\end{figure*}
\textbf{3D object detection on the InScope-Pri dataset}: The adopted baseline follows the convention \cite{zhang2022not,qian20223d,yang20203dssd,yan2018second}. \textbf{Point}-based detectors (such as PointRCNN \cite{shi2019pointrcnn}) directly extract spatial structure features from the raw point cloud data to obtain object proposals. However, the enormous quantity of raw data introduces a substantial computational burden, seriously impacting the computational efficiency. \textbf{Voxel}-based detectors (such as Pointpillar \cite{lang2019pointpillars}) can improve computational efficiency by converting point clouds into regular voxel grids. However, this discretization may compromise the detection accuracy, particularly for objects of varying sizes. \textbf{voxel and point}-based (such as PV-RCNN++ \cite{shi2020pv,shi2023pv}) detectors can integrate the advantages of point-based and voxel-based methods, allowing the detectors to fully use the spatial structure characteristics of the raw data while reducing their dependence on hardware. In addition, the popular \textbf{center}-based detectors (such as CenterPoint \cite{yin2021center,liu2023centertube}) adopt novel detection heads that can regress the an object's center point and orientation. 

\textbf{3D object detection on the InScope dataset}: The same methods are adopted here to analyze the effectiveness of the method on the fusion data that have more complete feature expressions, including the PointRCNN, Pointpillar, PV-RCNN++, and CenterPoint detectors. These detection performances can serve as the control group experiment to analyze the anti-occlusion capabilities.\par

\textbf{Anti-occlusion capability}: This section introduces a benchmark that provides consistent standards for evaluating and comparing the anti-occlusion capabilities of different methods, using $\xi_{D}$ as an evaluation metric. Specifically, this involves two datasets. The first is the InScope-Pri dataset, which contains partially and potentially occluded data. The second is the InScope dataset, a fusion dataset with relatively complete data. The disparities in the performance gaps and mean detection performance degradation ratio between the two datasets reflect the anti-occlusion capabilities of various strategies. More details about how to quantitatively measure the anti-occlusion capabilities of different methods can be found in Section \ref{detectionAnalysis}.

\begin{table*}
  \centering
  \caption{The collaborative perception performance gap ratio of each class, $mAP$, and mean collaborative perception performance degradation ratio $\xi_{D}$ (in \%). }
  \footnotesize
    \begin{tabular}{cccccc|cc|c}
    \toprule
    Paradigms & Methods & $G_{car}$ ↑& $G_{pedestrain}$ ↑ & $G_{cyclist}$ ↑ & $G_{truck}$ ↑ & $G_{mclass}$ ↑ &$ G_{mAP}$ ↑ & $\xi_{D}$ ↓ \\
    \midrule
    {Point-based} & PointRCNN & 85.21  & \textbf{90.99}  & 77.83  & 93.97  & 87.00  & 87.76  & 23.65  \\
    {Voxel-based} & Pointpillar & 86.29  & 67.40  & 74.43  & \textbf{95.55}  & 80.92  & 84.52  & 31.61 \\
     {Voxel \& Point-based}    & PV-RCNN++ & \textbf{90.12}  & 84.90  & \textbf{86.31}  & 94.89  & \textbf{89.05}  & \textbf{89.82}  & \textbf{20.01} \\
    {Center-based} & CenterPoint & 79.38  & 70.43  & 70.55  & 85.33  & 76.42  & 77.12  & 41.06 \\
    \bottomrule
    \end{tabular}
  \label{tab:detectiongap}
\end{table*}

\begin{figure*}
  \centering{\includegraphics[scale=1]{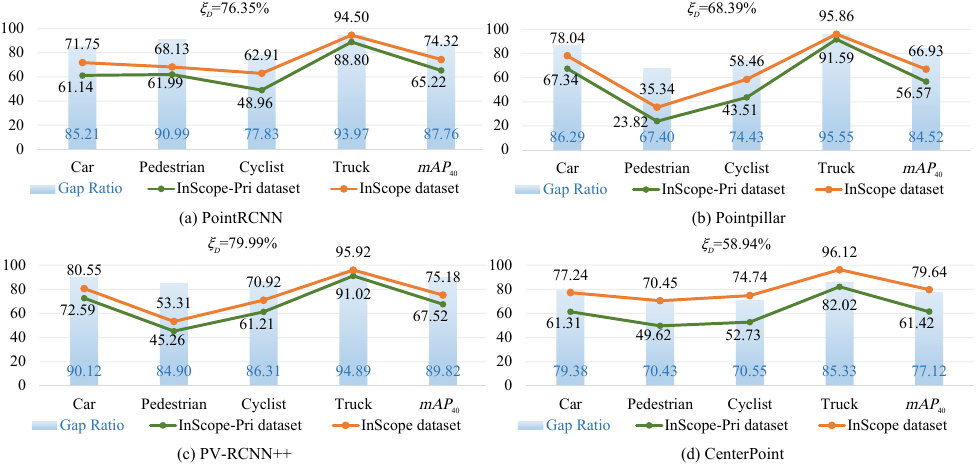}}
  \caption{Detection results on the InScope-Pri and InScope datasets and $\xi_{D}$ metric.}
  \label{fig:fusion_gap}
\end{figure*}
\subsubsection{Analysis}
\label{detectionAnalysis}
Here, the split ratios of the training/validation/test sets of the InScope dataset are set to 0.63:0.27:0.1. The 3D detection baselines are reproduced on the InScope-Pri and InScope datasets. 

\textbf{3D Object detection on the InScope-Pri and InScope datasets}: 
The detection results of different methods on the InScope-Pri and InScope validation sets are indicated in Table \ref{tab:detction_inscope}. The bottom of Table \ref{tab:detction_inscope} can be summarized as follows: \textbf{trucks} have extremely discriminative features, which improve significantly the detection performance in trucks. As the size of the objects decreases (such as in \textbf{car}, \textbf{cyclist}, and \textbf{pedestrian}), the detection performance of point-based and voxel-based methods decreases sharply. For instance, Table \ref{tab:detction_inscope} shows that the detection accuracy of PointRCNN and Pointpillar in the cyclist class are only 62.91\% and 58.46\%, respectively. The detection performance of the voxel and point-based detectors changes relatively slowly, and the detection accuracy of PV-RCNN++ in the cyclist class can still reach 70.92\%. Notably, benefiting from the mechanism of the center-based method of returning object properties, the detection accuracy of small objects such as pedestrians and cyclists improves dramatically. For example, CenterPoint can still achieve 74.74\% in the cyclist class, respectively. Meanwhile, Figure \ref{fig:detection_compare} indicates the detection results of the CenterPoint method on the InScope dataset. As seen from Figure \ref{fig:detection_compare}, the CenterPoint method based on the InScope dataset has a higher object detection performance than that based on the InScope-Pri dataset.

\textbf{3D Object detection on the InScope-Pri and InScope datasets}: more specific experimental results to to quantitatively analyze the anti-occlusion capabilities of various methods are presented in Table \ref{tab:detectiongap} and Figure \ref{fig:fusion_gap}. Generally, methods are expected to exhibit superior performance on the InScope dataset due to the richer point cloud representations than on the InScope-Pri dataset. To quantitatively measure and assess the anti-occlusion detection capabilities of different methodologies, more metrics, i.e., $G_{mclass}$, $ G_{mAP}$, and $\xi_{D}$ are adopted in Table \ref{tab:detectiongap}.
It follows that the smaller the degradation ratio $\xi_{D}$ is on the two datasets, the better the anti-occlusion ability of the methods, and vice versa. As seen from Figure \ref{fig:fusion_gap}, the PV-RCNN++ exhibits lower performance degradation ratio (79.99\%) than Pointpillar and CenterPoint, highlighting the superiority of the anti-occlusion capabilities and overall robust performance achieved by integrating the point and voxel features.

\subsection{Multisource Data Fusion}

\subsubsection{Metrics} Similar to the 3D object detection benchmark, the $AP_{40}$ of each class and $mAP_{40}$ are adopted here to evaluate the performance of the multisource data fusion task.

\subsubsection{Baselines} The three most commonly discussed fusion mechanisms are early fusion, middle fusion, and late fusion \cite{wang2023multi,zhang2023spatiotemporal}. Their characteristics and disadvantages are discussed and reported as follows:

$\bullet$ The \textbf{early fusion} mechanism directly combines one frame of the raw data pair ${\bf{X}} = \left[ {{\bf{X}}_p,{\bf{X}}_s} \right]$ captured by the InScope-Pri dataset ${{\bf{X}}_p}$ and the InScope-Sec dataset ${\bf{X}}_s$. Then, ${\bf{X}}$ is fed into 3D detectors to obtain the final proposals.

$\bullet$ The \textbf{late fusion} adopts two independent 3D detectors to obtain two sets of prediction proposals. Subsequently, post-optimization operations such as voting or weight summarization are performed to obtain the final proposals.

\begin{table*}
 \caption{Comparative results of different fusion mechanisms based on the PV-RCNN++ backbone. The best results under different fusion mechanisms are marked in \textbf{bold}. }
  \centering
  \footnotesize
    \begin{tabular}{ccccccccc}
    \toprule
   \multicolumn{1}{c}{Detector} & \multicolumn{2}{c}{Fusion Mechanism} & \multicolumn{1}{m{6em}<{\centering}}{\ \ Car ↑\newline{}$AP_{40(IoU=0.7)}$} & \multicolumn{1}{m{6em}<{\centering}}{Pedestrian ↑\newline{}$AP_{40(IoU=0.5)}$} & \multicolumn{1}{m{6em}<{\centering}}{Cyclist ↑\newline{}$AP_{40(IoU=0.5)}$} & \multicolumn{1}{m{6em}<{\centering}}{\  Truck ↑\newline{}$AP_{40(IoU=0.7)}$} & $mAP_{40}$ ↑ & FPS ↑\\
    \midrule
   \multirow{5}[0]{*}{PV-RCNN++} & \multirow{2}[0]{*}{No Fusion} & InScope-Sec  & 43.49 & 34.60 & 39.94 & 76.04 &  48.52  & \textbf{16.67} \\
                            &      & InScope-Pri  & 72.59 & 45.26 & 61.21 & 91.02 & 67.52  & 13.28  \\
    \cmidrule{2-9}
    &\multicolumn{2}{c}{Early Fusion (InScope)} & \textbf{80.55} &  53.31 & \textbf{70.92} & \textbf{95.92} & \textbf{75.18} & 14.66  \\
    &\multicolumn{2}{c}{Late Fusion} & 68.01 &  \textbf{53.47} & 56.95 &92.65 & 67.77  & 1.21  \\
    &\multicolumn{2}{c}{Middle Fusion} & 73.78 &  52.06 & 62.06 & 91.89  & 69.95  & 13.02  \\
    \bottomrule
    \end{tabular}
  \label{tab:fusionresult_pvrcnn++}
\end{table*}

\begin{table*}
 \caption{Comparative results of different fusion mechanisms based on the CenterPoint backbone. The best results under different fusion mechanisms are marked in \textbf{bold}. }
  \centering
  \footnotesize
    \begin{tabular}{ccccccccc}
    \toprule
   \multicolumn{1}{c}{Detector} & \multicolumn{2}{c}{Fusion Mechanism} & \multicolumn{1}{m{6em}<{\centering}}{\ \ Car ↑\newline{}$AP_{40(IoU=0.7)}$} & \multicolumn{1}{m{6em}<{\centering}}{Pedestrian ↑\newline{}$AP_{40(IoU=0.5)}$} & \multicolumn{1}{m{6em}<{\centering}}{Cyclist ↑\newline{}$AP_{40(IoU=0.5)}$} & \multicolumn{1}{m{6em}<{\centering}}{\  Truck ↑\newline{}$AP_{40(IoU=0.7)}$} & $mAP_{40}$ ↑ & FPS ↑\\
    \midrule
    \multirow{5}[0]{*}{CenterPoint} &\multirow{2}[0]{*}{No Fusion} & InScope-Sec  & 35.92 & 37.40 & 38.24 & 68.78 & 45.08  & \textbf{107.53} \\
                        &          & InScope-Pri  & 61.31 & 49.62 & 52.73 & 82.02 & 61.42  & 31.45  \\
    \cmidrule{2-9}
    &\multicolumn{2}{c}{Early Fusion (InScope)} & \textbf{77.24} & \textbf{70.45} & \textbf{74.74} & \textbf{96.12} &  \textbf{79.64} & 30.49  \\
    &\multicolumn{2}{c}{Late Fusion}   & 58.13  & 50.03  & 56.01  & 85.65  &  62.45  & 6.40  \\
    &\multicolumn{2}{c}{Middle Fusion} & 52.74  & 38.95  & 51.19  & 81.73  &  56.15  & 15.85  \\
    \bottomrule
    \end{tabular}
  \label{tab:fusionresult_centerpoint}
\end{table*}

$\bullet$ The \textbf{middle fusion} focuses on the spatial feature representations $\left( {{\bf{F}}_p,{\bf{F}}_s} \right)$ from different models. Then, a fusion module based on the convolution neural network or transformer is adapted to align the extracted features and generate the final proposals. 

Inspired by the adaptive feature fusion module \cite{wang2022am3net}, a simple and efficient fusion baseline is adopted for this benchmark, as shown in Figure \ref{fig:middlefusion}. The adopted fusion module fuses the point cloud feature representations extracted by the VoxelNet backbone \cite{zhou2018voxelnet}.

\subsubsection{Analysis} In this experiment, PV-RCNN++ and CenterPoint, which serve as typical models in 3D detection paradigms, are adopted as baselines for exploring the advantages of different fusion mechanisms. The experimental results are shown in Table \ref{tab:fusionresult_pvrcnn++} and Table \ref{tab:fusionresult_centerpoint}. As seen, for the PV-RCNN++ method, the top two lines report the experimental results using single raw data captured from the InScope-Sec dataset ${{\bf{X}}_s}$ and the InScope-Pri dataset ${{\bf{X}}_p}$, respectively. It can be found from Table \ref{tab:fusionresult_pvrcnn++} that the dense point cloud captured from ${{\bf{X}}_p}$ can provide more sufficient information than the sparse point cloud from ${{\bf{X}}_s}$. Moreover, the computational efficiency is negatively affected. For example, the $mAP_{40}$ of PV-RCNN++ can obtain 48.52\% and 67.52\% with the secondary point cloud and principal point cloud, respectively, while the FPS decreased from 16.67 to 13.81.

\begin{figure}
  \centering{
  \includegraphics[scale=1]{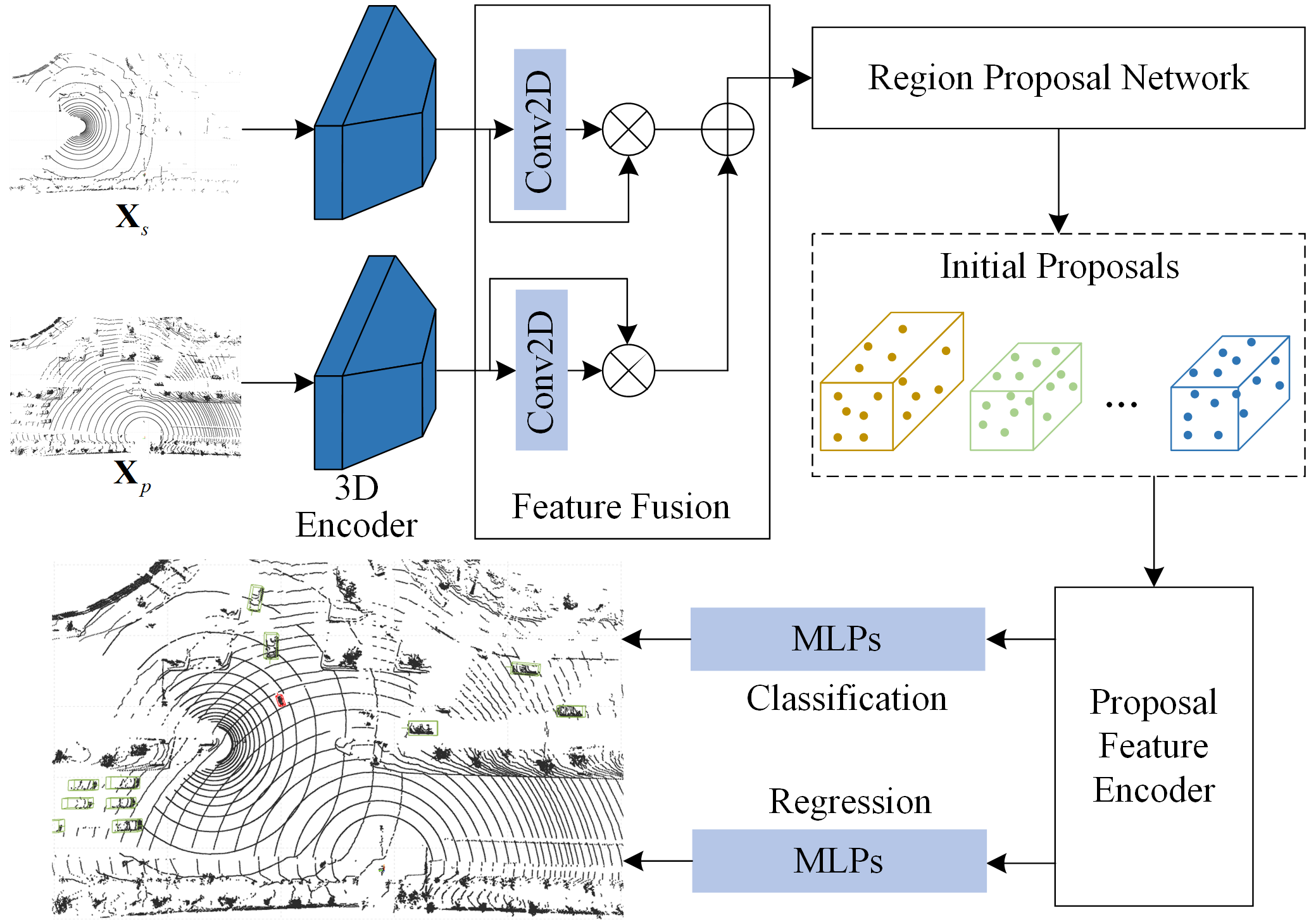}}
  \caption{An illustration of the middle fusion module.}
  \label{fig:middlefusion}
\end{figure}

\begin{figure*}
  \centering{
  \includegraphics[scale=1]{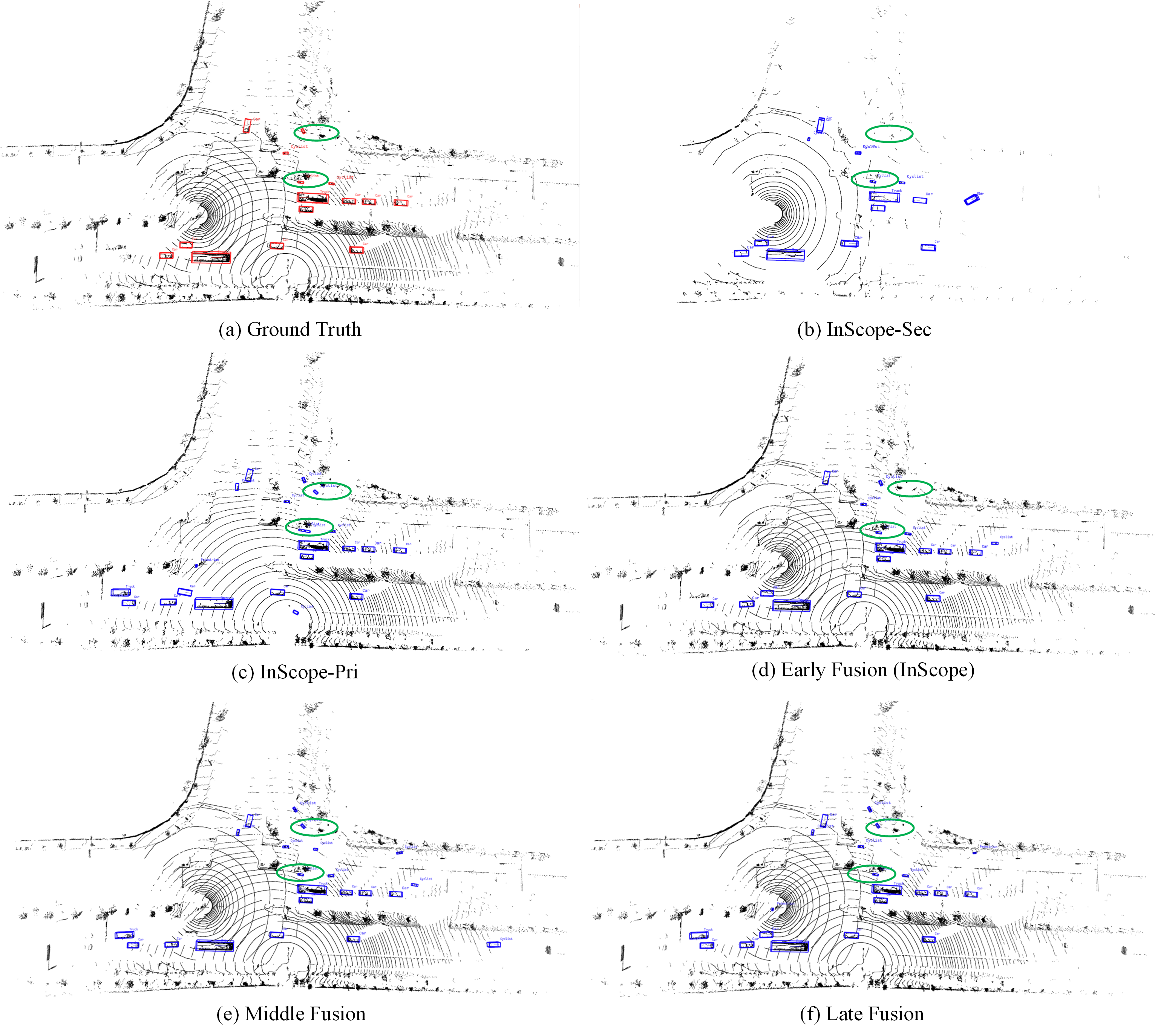}}
  \caption{The detection results of the PV-RCNN++ detector are based on the different fusion mechanisms on frame 020773. The red and blue bounding boxes represent the ground truth and detection results, respectively. The green circles represent the differences among them.}
  \label{fig:fusion_detection}
\end{figure*}

More importantly, the data fusion strategy increases point cloud density and completeness. The detection performances of different types of fusion strategies, such as early, middle, and late fusions, its detection performances show varying degrees of growth compared with those of single raw data. Specifically, early fusion is a straightforward data fusion mechanism; it exhibits the most significant improvement in detection performance ($mAP_{40}$ = 75.18\%). The improvement in the detection performance is relatively limited with the late-fusion mechanism($mAP_{40}$ = 67.77\%). Moreover, designing various middle fusion architectures focused on feature-level fusion has emerged as a key research direction in multisource data fusion. Here, a simple but effective middle fusion strategy is provided in Figure \ref{fig:middlefusion}, which achieved specific results ($mAP_{40}$ = 69.95\%); however, considerable opportunities for improvement remain. Figure \ref{fig:fusion_detection} shows the detection performance of different fusion strategies.

Another set of experiments based on the CenterPoint backbone is conducted to analyze the effectiveness of the fusion mechanism for different detection paradigms, as shown in Table \ref{tab:fusionresult_centerpoint}. Unlike that of PV-RCNN++ backbone, the middle fusion detection performance of CenterPoint is slightly inferior to that of the late fusion mechanism (56.15\% < 62.45\%). This is because the detection head based on the center heatmap head designed by CenterPoint adopts a new idea for locating the object by calculating the probability that each voxel belongs to the center point \cite{zhou2020tracking}. The simple fusion module designed considers only the integration of forwards features and does not comprehensively consider the impact of different detection heads on backpropagation. Therefore, researchers can attempt more intelligent middle fusion modules by comprehensively considering the implications between the detection backbone and detection head.

\begin{table*}
   \caption{Data domain transfer results of different unsupervised domain adaptation methods on \textbf{car} class. ``Source→Target'' represents that methods are pre-trained on the source domain and transferred to the target domain.}
  \centering
  \footnotesize
    \begin{tabular}{cccccccccc}
    \toprule
    \multirow{2}[0]{*}{Source→Target} &\multicolumn{2}{c}{DAIR-V2X-I→KITTI} & \multicolumn{2}{c}{ONCE→KITTI} & \multicolumn{2}{c}{InScope→KITTI} &  \multicolumn{2}{c}{InScope→DAIR-V2X-I} & DAIR-V2X-I→InScope \\
    & Moderate & $mAP_{40}$ & Moderate & $mAP_{40}$ & Moderate & $mAP_{40}$ & Moderate & $mAP_{40}$ & $AP_{40(IoU=0.7)}$ \\
    \midrule
    Source Domain &    36.47 & 37.98 & 38.65 & 41.65 &49.45 & 52.97 &  29.54 & 31.05 & 32.16 \\
    SN            &    44.76 & 44.80 & 45.95 & 49.34 &58.66 & 61.87 &  30.47 & 31.81 & 33.25 \\
    ST3D          &    62.04 & 65.35 & 53.92 & 58.19 &70.06 & 74.63 &  34.65 & 48.98 & 37.03 \\
    Target Domain &    78.63 & 81.63 & 78.63 & 81.63 &78.63 & 81.63 &  78.51 & 81.41 & 71.75 \\
    \bottomrule
    \end{tabular}
   \label{tab:transfer}%
\end{table*}%

\subsection{Data Domain Transfer}
\subsubsection{Metrics} The $AP_{40}$ metric is employed to assess the performance of unsupervised domain adaptation methods on the InScope dataset. The $mAP_{40}$ based on the easy, moderate, and hard difficulty levels is utilized to evaluate the adaptability of other datasets.

\begin{figure}
  \centering{
  \includegraphics[scale=0.98]{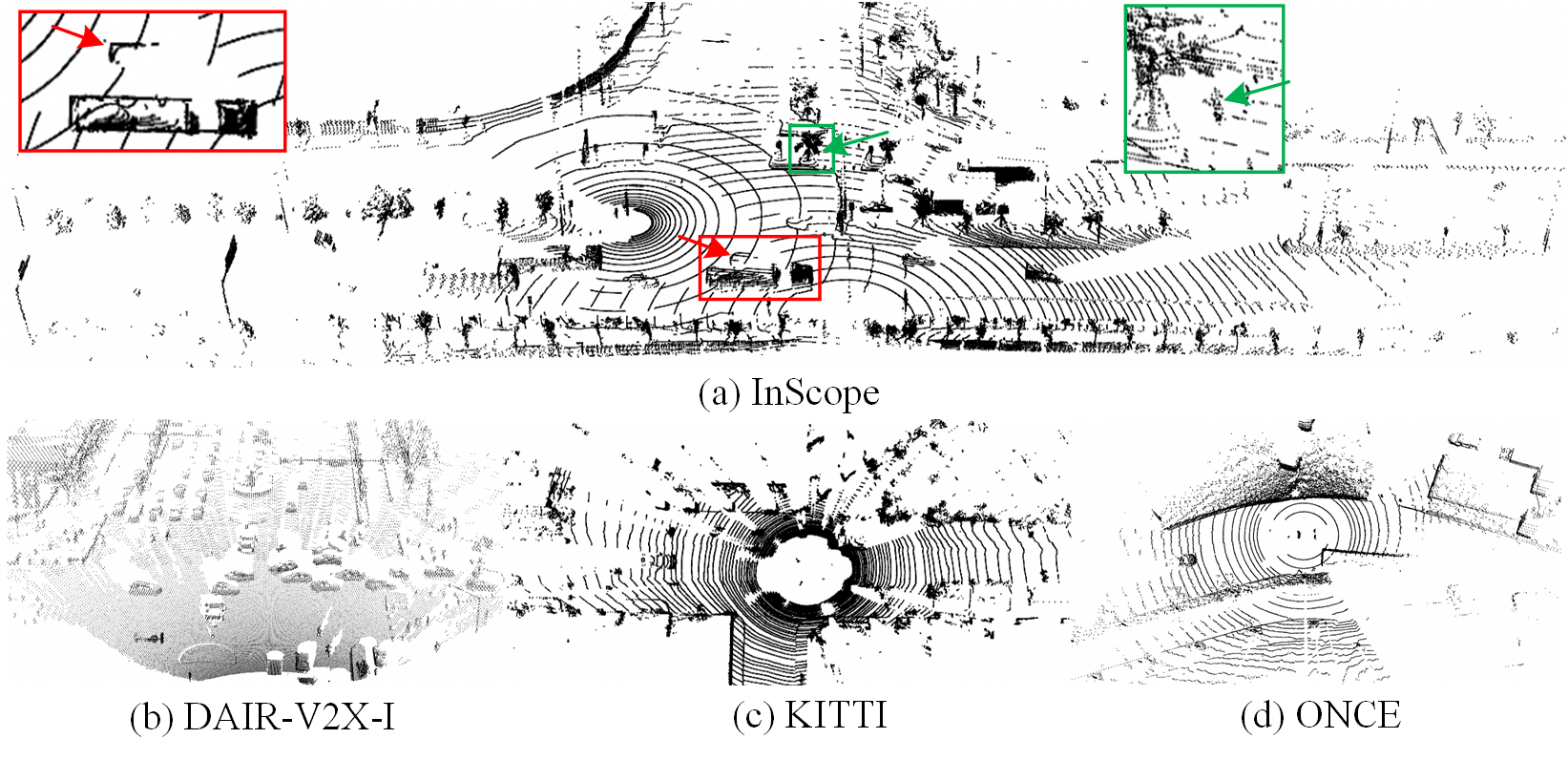}}
  \caption{Point clouds of the InScope, DAIR-V2X-I, KITTI, and ONCE datasets.}
  \label{fig:domain}
\end{figure}
\subsubsection{Baselines} In this analysis, the point cloud data from four domains are utilized for data domain transfer analysis, i.e., vehicle-side-based KITTI and ONCE and infrastructure-side-based DAIR-V2X-I and InScope. Their scene visualizations are shown in Figure \ref{fig:domain}. It can be found that four different datasets have distinct point cloud density attribute characteristics. Therefore, this benchmark is used to validate the efficacy of different datasets in mitigating \textbf{domain gaps}. Notably, the InScope, KITTI, and ONCE datasets are generated by the mechanical LiDAR, whereas the DAIR-V2X-I dataset is captured by solid-state LiDAR. As a result, a notable disparity exists in the spatial point cloud distribution. In addition to exploring the domain gap, this experiment discusses the \textbf{scan pattern gap} for 3D detector capability in solid-state and mechanical LiDAR. Here, the PointRCNN is used as the 3D detection backbone; the experimental results with different unsupervised transfer methods are reported in Table \ref{tab:transfer}. 

Specifically, ``Source Domain'' indicates that the PointRCNN is trained on the source data domain and tested on the target data domain. ``Target Domain'' indicates that the PointRCNN is trained and tested on the target data domain. In addition, two widely used unsupervised data domain transfer learning methods are adopted, including statistical normalization (SN) \cite{wang2020train} and self-training 3D object detection (ST3D) \cite{yang2021st3d}. 

\subsubsection{Analysis} For the \textbf{domain gap}, it can be seen from Table \ref{tab:transfer} that the detection performance is improved significantly when the PointRCNN pre-trained on the InScope dataset is transferred to the KITTI dataset. For example, after optimization by the ST3D method, the $mAP_{40}$ can improve from 52.97\% to 74.63\% in the KITTI dataset. Its detection performance is closer to that of methods trained on the KITTI dataset ($mAP_{40}$ = 81.63\%) than pre-training in the DAIR-V2X-I ($mAP_{40}$ = 65.35\%) and ONCE ($mAP_{40}$ = 58.19\%). It is proven that the InScope dataset can solve the domain gap better than other datasets. Moreover, the results demonstrate that the proposed InScope dataset is more conducive to transferring 3D detectors to other data domains. 

\begin{figure}
  \centering{
  \includegraphics[scale=1]{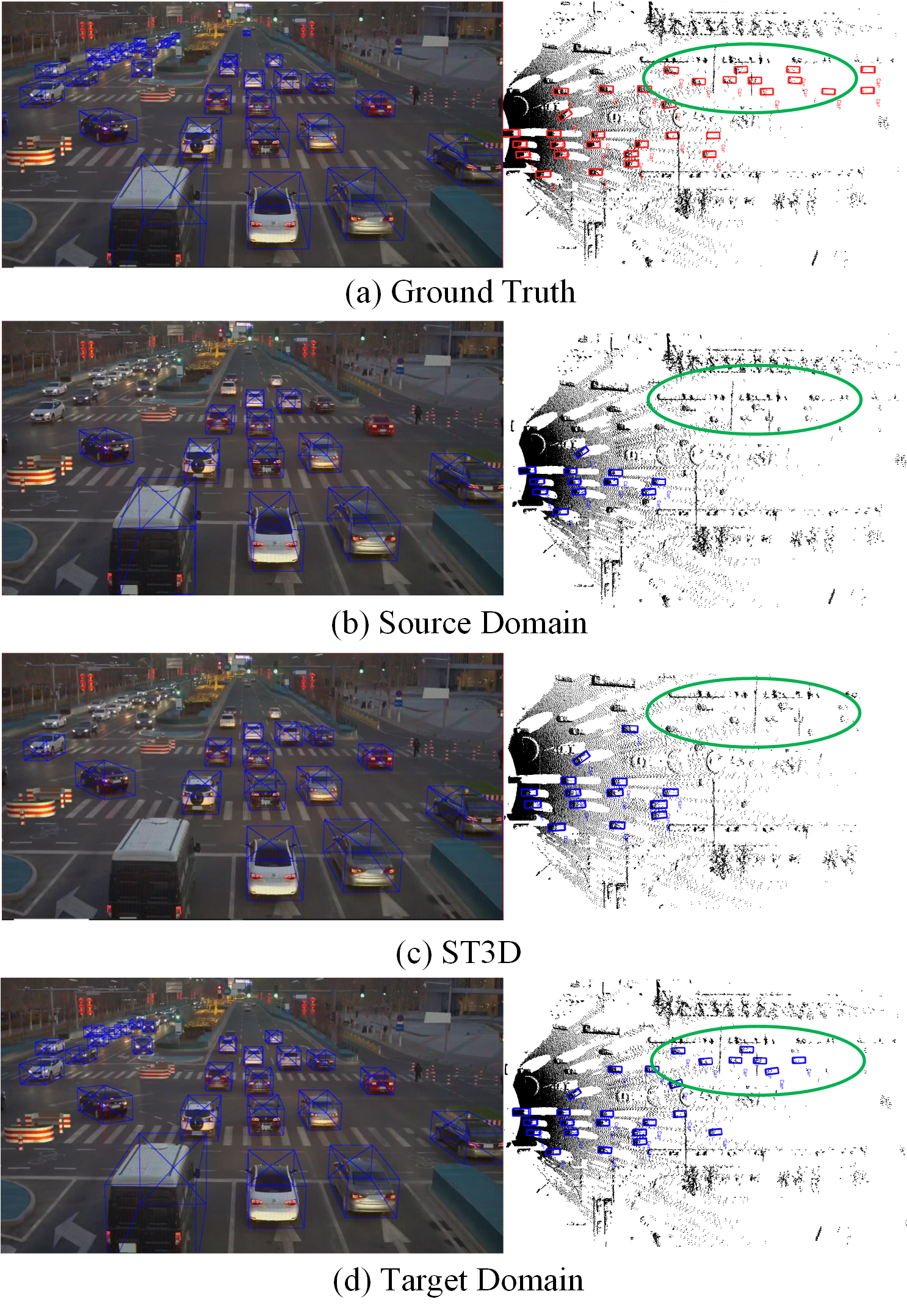}}
  \caption{The data domain transfer results (InScope → DAIR-V2X-I) based on the PointRCNN backbone. The green circles represent the differences among them.}
  \label{fig:domaintransfer}
\end{figure}

However, for the \textbf{scan pattern gap}, the performance improvement is not ideal when the pre-trained PointRCNN in the InScope dataset is transferred to the DAIR-V2X-I dataset. For example, the $mAP_{40}$ can only be improved from 31.05\% to 48.98\%. This demonstrates the challenges inherent in the domain adaptation of 3D detectors across datasets captured by the scan pattern gap of LiDAR. Figure \ref{fig:domaintransfer} shows the experimental results that transfer the pre-trained detection methods from the InScope dataset to the DAIR-V2X-I datasets under the domain and scan patterns gap. This proves that the scan pattern gap in LiDAR technology still requires further exploration.

\begin{table*}
  \centering
    \caption{Tracking results of the AB3DMOT on the \textbf{car} class. The best results under each $IoU_{thres}$ are marked in \textbf{bold}.}
   \footnotesize
    \begin{tabular}{cccccccccc}
    \toprule
    \multirow{2}[0]{*}{Tracker} & \multirow{2}[0]{*}{Detectors} & \multicolumn{4}{c}{Car ($IoU_{thres}=0.5$)}   & \multicolumn{4}{c}{Car ($IoU_{thres}=0.7$)} \\
\cmidrule{3-10}          &       & sAMOTA ↑ & AMOTA ↑ & MOTA ↑ & FRAG ↓ & sAMOTA ↑ & AMOTA ↑ & MOTA ↑ & FRAG ↓ \\
    \midrule
    \multirow{4}[0]{*}{AB3DMOT} & PointRCNN & 74.81  & 30.63  & 63.25  & 595   & 60.34  & 20.82  & 44.45  & 1834 \\
          & Pointpillar & \textbf{82.23} & 37.44  & \textbf{68.85} & 391   & 64.98  & 24.56  & 46.82  & 2166 \\
          & PV-RCNN++ & 81.63  & \textbf{36.97} & 67.56  & 386   & \textbf{68.71} & \textbf{26.96} & \textbf{50.72} & \textbf{1560} \\
          & CenterPoint & 78.76  & 35.01  & 61.02  & \textbf{367} & 61.25  & 22.34  & 40.98  & 1720 \\
    \bottomrule
    \end{tabular}
  \label{tab:tracking_car}
\end{table*}

\begin{table*}
  \centering
   \caption{Tracking results of the AB3DMOT on the \textbf{pedestrian} class. The best results under each $IoU_{thres}$ are marked in \textbf{bold}.}
   \footnotesize
    \begin{tabular}{cccccccccc}
    \toprule
    \multirow{2}[0]{*}{Tracker} & \multirow{2}[0]{*}{Detectors} & \multicolumn{4}{c}{Pedestrian ($IoU_{thres}=0.25$)} & \multicolumn{4}{c}{Pedestrian ($IoU_{thres}=0.5$)} \\
\cmidrule{3-10}          &       & sAMOTA ↑ & AMOTA ↑ & MOTA ↑ & FRAG ↓ & sAMOTA ↑ & AMOTA ↑ & MOTA ↑ & FRAG ↓ \\
    \midrule
    \multirow{4}[0]{*}{AB3DMOT} & PointRCNN & 59.89  & 21.24  & 39.73  & 6     & 56.59  & 18.78  & 37.06  & 22 \\
          & Pointpillar & 32.09  & 2.85  & 27.79  & \textbf{4} & 27.42  & 1.83  & 25.36  & 24 \\
          & PV-RCNN++ & 31.39  & 10.73  & 27.71  & 10    & 28.54  & 8.76  & 25.75  & \textbf{20} \\
          & CenterPoint & \textbf{67.38} & \textbf{28.23} & \textbf{63.48} & 8     & \textbf{62.03} & \textbf{24.94} & \textbf{59.30} & 35 \\
    \bottomrule
    \end{tabular}
  \label{tab:tracking_pedestrain}
\end{table*}

\begin{table*}
  \centering
   \caption{Tracking results of the AB3DMOT on the \textbf{cyclist} class. The best results under each $IoU_{thres}$ are marked in \textbf{bold}.}
   \footnotesize
    \begin{tabular}{cccccccccc}
    \toprule
    \multirow{2}[0]{*}{Tracker} & \multirow{2}[0]{*}{Detectors} & \multicolumn{4}{c}{Cyclist ($IoU_{thres}=0.25$)} & \multicolumn{4}{c}{Cyclist ($IoU_{thres}=0.5$)} \\
\cmidrule{3-10}          &       & sAMOTA ↑ & AMOTA ↑ & MOTA ↑ & FRAG ↓ & sAMOTA ↑ & AMOTA ↑ & MOTA ↑ & FRAG ↓ \\
    \midrule
    \multirow{4}[0]{*}{AB3DMOT} & PointRCNN & 60.97  & 22.75  & 41.56  & 99    & 50.27  & 15.42  & 33.77  & 272 \\
          & Pointpillar & 49.96  & 19.18  & 33.82  & \textbf{64} & 33.75  & 9.42  & 22.33  & 379 \\
          & PV-RCNN++ & 63.00  & 26.93  & 43.22  & 177   & 52.65  & 19.01  & 34.12  & 349 \\
          & CenterPoint & \textbf{68.78} & \textbf{29.07} & \textbf{45.42} & 70    & \textbf{57.50}  & \textbf{20.80} & \textbf{37.58}  & \textbf{267} \\
    \bottomrule
    \end{tabular}
  \label{tab:tracking_cyclist}
\end{table*}

\subsection{3D Multiobject Tracking}
\label{tracking}

\subsubsection{Metrics} Two evaluation metrics are used, i.e., sAMOTA, AMOTA, and CLEAR metrics (including MOTA and FRAG) \cite{bernardin2008evaluating, Weng2020_AB3DMOT}. For object matching criteria, 3D IoU is used to judge whether interframe objects are successfully associated \cite{zheng2020distance,huang2021joint}. Two 3D IoU thresholds ($IoU_{thres}$) for each class are reported in the 3D multiobject tracking benchmark, to evaluate the tracking performance of the methods under various difficulties, which is different from the IoU setting of 3D object detection \cite{liu2023centertube}.

\subsubsection{Baselines} The 3D multiobject tracking methods can be divided into two paradigms \cite{9999157,ijcai2023p143,chen2023voxelnext,wu20213d}: joint detection and tracking (JDT) \cite{zhang2023spatiotemporal,huang2021joint,wang2021ditnet} and tracking-by-detection (TBD) \cite{kim2022polarmot,sun2021you,10304295,pang2023standing,marinello2022triplettrack}. Due to the lower computational efficiency of the JDT-based methods, TBD-based methods are used as baselines in this paper, i.e., AB3DMOT \cite{Weng2020_AB3DMOT}. For the 3D multiobject tracking, the tracking results are reported for all sequences. The pre-trained PointRCNN, Pointpillar, PV-RCNN++, and CenterPoint are used as detectors to obtain the detection status of the object at time $t$. The 3D IoU is used as a matching criterion \cite{zheng2020distance} to associate the detected objects at time $t$ with the tracking objects at time $t$-1,  If the 3D IoU of objects between interframes is greater than the threshold ($IOU_{thres}$), the objects can be successfully matched, and vice versa. To evaluate the tracking performance under different difficulties in the InScope dataset, the $IOU_{thres}$ of [car, truck, pedestrian, cyclist] are set to $\left[\left[0.5 / 0.7\right], \left[0.5 / 0.7\right], \left[0.25 / 0.5\right], \left[0.25 / 0.5\right]\right]$.

\begin{table*}
  \centering
   \caption{Tracking results of the AB3DMOT on the \textbf{truck} class. The best results under each $IoU_{thres}$ are marked in \textbf{bold}.}
   \footnotesize
    \begin{tabular}{cccccccccc}
    \toprule
    \multirow{2}[0]{*}{Tracker} & \multirow{2}[0]{*}{Detectors} & \multicolumn{4}{c}{Truck ($IoU_{thres}=0.5$)}   & \multicolumn{4}{c}{Truck ($IoU_{thres}=0.7$)} \\
\cmidrule{3-10}          &       & sAMOTA ↑ & AMOTA ↑ & MOTA ↑ & FRAG ↓ & sAMOTA ↑ & AMOTA ↑ & MOTA ↑ & FRAG ↓ \\
    \midrule
    \multirow{4}[0]{*}{AB3DMOT} & PointRCNN & \textbf{82.53} & 36.33  & 73.34  & 124   & \textbf{78.67} & 33.49  & 68.20  & 181 \\
          & Pointpillar & 82.18  & \textbf{37.20} & \textbf{75.26} & 80    & 76.79  & \textbf{33.88} & \textbf{70.33} & 182 \\
          & PV-RCNN++ & 81.50  & 36.35  & 69.15  & 76    & 77.20  & 33.41  & 64.53  & \textbf{141} \\
          & CenterPoint & 81.44  & 36.58  & 71.89  & \textbf{70} & 76.11  & 33.12  & 65.85  & 207 \\
    \bottomrule
    \end{tabular}
  \label{tab:tracking_truck}
\end{table*}

\subsubsection{Analysis} The tracking performances for cars, pedestrians, cyclists, and trucks are shown in Table \ref{tab:tracking_car}, Table \ref{tab:tracking_pedestrain}, Table \ref{tab:tracking_cyclist}, and Table \ref{tab:tracking_truck}, respectively. The detector performance significantly affects the tracking performance of the TBD-based tracking method (AB3DMOT). Meanwhile, the larger the object volume is, the more sufficient the matching criteria provided for the tracking module. As seen from Table \ref{tab:tracking_car}, for the \textbf{car} class, the sAMOTA metric of the AB3DMOT method based on the pre-trained CenterPoint has 82.23\% when the $IOU_{thres}$ is set to 0.5. However, when $IOU_{thres}$ increases from 0.5 to 0.7, the sAMOTA metric decreases sharply from 82.23\% to 64.98\%, indicating that the proposed InScope dataset still presents significant challenges in a higher 3D IoU tracking task. These phenomenon are even more pronounced for smaller object (i.e., pedestrians and cyclists) tracking, as shown in Table \ref{tab:tracking_pedestrain} and Table \ref{tab:tracking_cyclist}. For instance, the tracking performance for the CenterPoint detector in the sAMOTA metric is only 57.50\% when the $IOU_{thres}$ is set to 0.5. More advanced detection and tracking frameworks are worth further design and exploration.

\begin{figure}
  \centering{
  \includegraphics[scale=0.82]{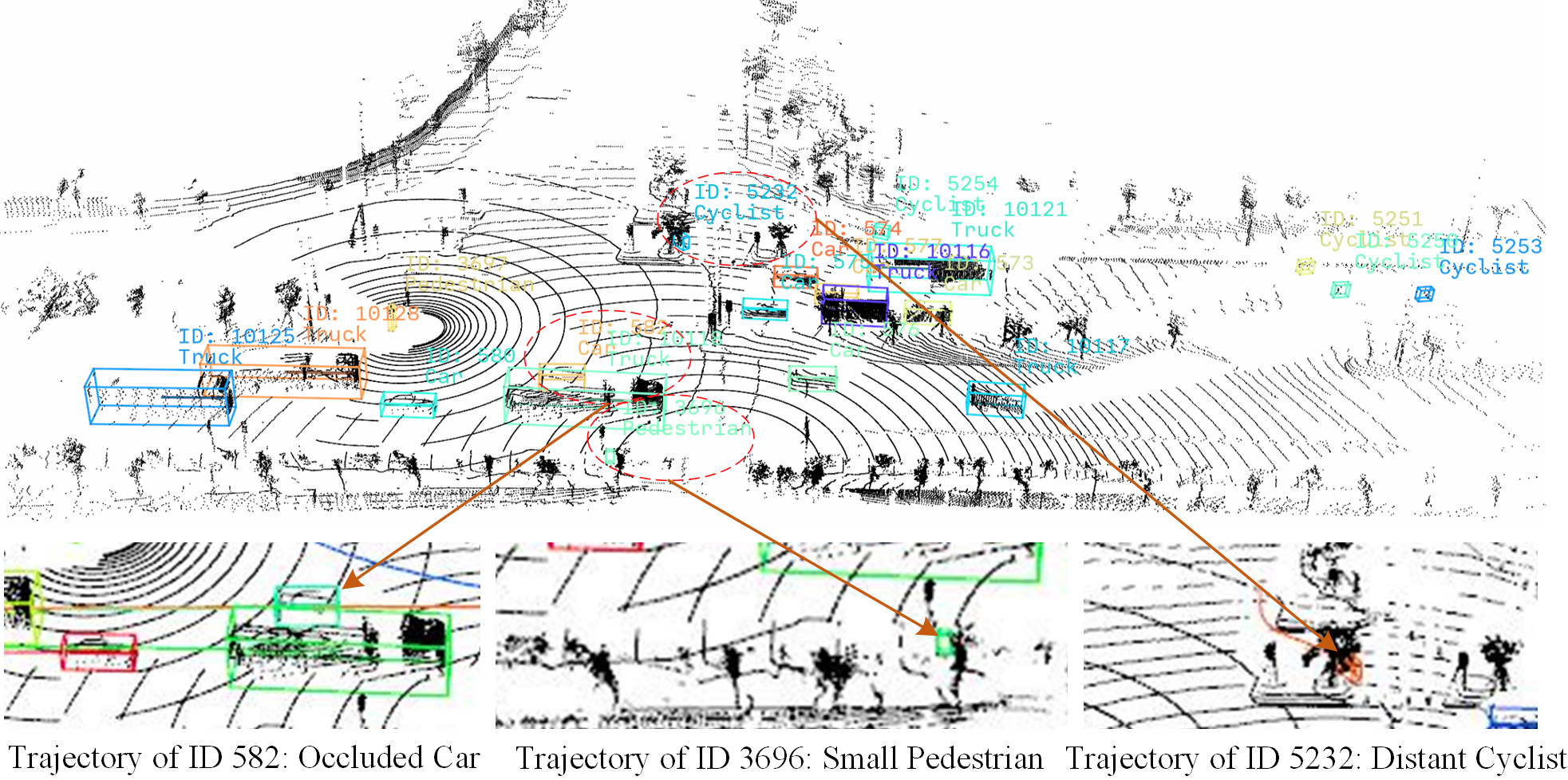}}
  \caption{Tracking results of the AB3DMOT method based on the CenterPoint detector on the sequence 21.}
  \label{fig:track}
\end{figure}

Finally, some tracking examples of the AB3DMOT based on the pre-trained CenterPoint method are shown in Figure \ref{fig:track}. The InScope dataset can accurately track occluded, small, or distant objects through supplementary data. Specifically, when an occluded car (ID 582) passes an intersection, the perception information of the InScope-Pri dataset is completely lost. However, ID 582 can still be stably and continuously tracked with the supplementary data of the InScope-Pri dataset. Moreover, a small pedestrian (ID 3696) and a distant occluded cyclist (ID 5232) can also be tracked stably and continuously. This proves that the scan pattern gap in LiDAR technology requires further exploration.

\section{CONCLUSION}
\label{conclusions}
This paper proposes an innovative infrastructure-side, real-world InScope dataset that performs I2I visual blind-spot compensation and quantitatively evaluates the anti-occlusion detection capabilities of various methods. Furthermore, the dataset establishes four critical perception benchmarks, advancing the standardization of perception models within the V2X community. In addition, comprehensive experimental evaluations utilizing widely used methods are conducted across four benchmarks, which sets a rigorous baseline for subsequent research endeavours. Experiments indicate that the InScope dataset can provide robust tracking performance for obscured and small objects at a distance or in occluded areas, which can significantly enhance perception systems in V2X, fostering the development of more reliable autonomous vehicular technologies.

\printcredits

\section*{Declaration of Competing Interest}
The authors declare that they have no known competing financial interests or personal relationships that could have appeared to influence the work reported in this paper.

\section*{Data availability}
The dataset and code are available at \url{https://github.com/xf-zh/InScope}.

\section*{Acknowledgment}
This work was supported in part by the Key-Area Research and Development Program of Guangdong Province under Grant 2020B0909030005 and in part by the Southern Marine Science and Engineering Guangdong Laboratory (Zhuhai) under Grant SML2020SP011. \par

\bibliographystyle{elsarticle-num}

\bibliography{refs}

\end{document}